# Multi-Robot System Architecture Design in SysML and BPMN

Ahmed R. Sadik[1,*], Christian Goerick[1]

[1]Honda Research Institute Europe, 63073, Germany

A B S T R A C T

*Multi-Robot System (MRS) is a complex system that contains many different software and hardware components. This main problem addressed in this article is the MRS design complexity. The proposed solution provides a modular modeling and simulation technique that is based on formal system engineering method, therefore the MRS design complexity is decomposed and reduced. Modeling the MRS has been achieved via two formal Architecture Description Languages (ADLs), which are Systems Modeling Language (SysML) and Business Process Model and Notation (BPMN), to design the system blueprints. By using those abstract design ADLs, the implementation of the project becomes technology agnostic. This allows to transfer the design concept from on programming language to another. During the simulation phase, a multi-agent environment is used to simulate the MRS blueprints. The simulation has been implemented in Java Agent Development (JADE) middleware. Therefore, its results can be used to analysis and verify the proposed MRS model in form of performance evaluation matrix.*

## 1. Introduction

This paper extends the work presented at the 2019 International Conference on Mechatronics, Robotics, and System Engineering (MoRSE) [1]. Related work can be also seen in [2].

Multi-Robot System (MRS) is a cyber-physical system that contains more than one robot, each of them owns a unique set of capabilities. The idea of an MRS is to solve a complex problem by collectively using the current capabilities of existing robots [3]. Therefore, the MRS must match the given problem with the existing robots' capabilities, to plan the solution steps. Many MRS applications can be seen in swarm robotics, cooperative automated transportation, unmanned aerial vehicles, and cooperative manufacturing [4]. The advantages of an MRS is increasing the performance by saving the time and the effort to solve the problem. Moreover, distributing the solution among different robots provides more computational processing power, this means faster and higher capacity to solve many problems simultaneously [5].

Implementing an MRS without a proper system architecture design is a crucial mistake that is often done by the system developers. Because the system requirements and functionalities are lost in a non-human readable machine code. Therefore, in this article we purpose a model driven development approach that uses the system model as the main software artifacts [6]. The proposed design approach in this article is based on the V-Model, which is a de facto solution for complex systems such as MRS.

*****
* Ahmed R. Sadik, ahmed.sadik@honda-ri.de

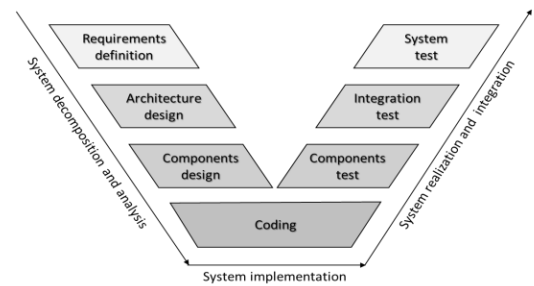

Figure 1: The V-Model simplified version – adapted from [6]

The V-Model shown in Figure 1 describes the required stages to build an MRS. In the first stage of the V-Model, the system is decomposed. In this stage, the system components and architecture are designed based on the system requirements. In the second stage, the implementation of the MRS is carried out. The implementation of an MRS often involves the coding the individual components. In the final stage, the MRS individual components are tested through unit tests, then integration tests are carried out over sub-systems and eventually the overall integrated system. This article focusses on the first stage of the V-Model to build an MRS. As the design stage is the most curtail stage of an MRS system building, because all the following stages are depending on this design.

Section 2 of the article describes the problem and the use case of concern. Section 3 introduces the background that that is needed to model and simulate the use case. Section 4 discusses the system requirements that are used to build and evaluate the system performance Modeling the use case is explained in detail in section 5, while its simulation is shown in section 6. Therefore, the performance analysis is explained in section 7. Ultimately, the last section concludes the work and the future research.

## 2. Problem and use case

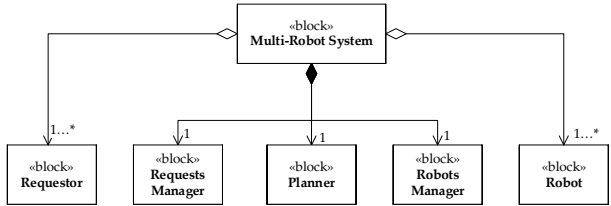

Figure 2: SysML block definition diagram for the proposed architecture

The main article objective is to design an MRS architecture model that can be simulated and evaluated due to a predefined evaluation criterion. An MRS architecture is an overall system description that abstracts its functionalities, logic, and constrains [7]. Accordingly, it provides an analysis tool to grasp and improve system characteristics, and a conceptual model that can be used as the system blueprints [8]. In this work, SysML block definition diagram is used to describe the proposed MRS architecture and components as shown in Figure 2 and Figure 3. SysML diagrams will be explained in the next section as many of them are used in constructing the proposed MRS design.

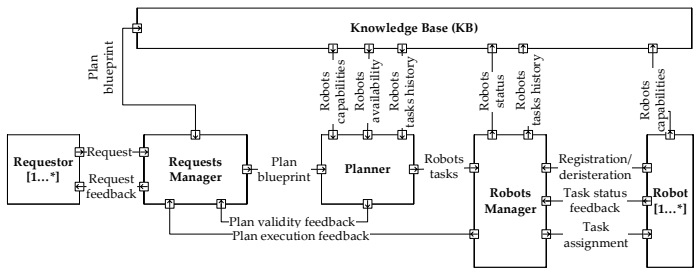

Figure 3: SysML internal block diagram for the proposed MRS architecture

Figure 2 is the the proposed MRS block definition diagram. The block definition diagram defines the main components of the architecture, which are the Requests Manager (RqM), the planner (PLN), and the Robots Manager (RbM). Figure 3 shows the proposed MRS internal block diagram that describes the connections among the components as illustrated in Figure 2. When the RqM receives a request (Rq), it checks if there is a plan-blueprint (Pb) in the Knowledge Base (KB) to fulfill this Rq. A Pb is a sequence of tasks (T), i.e., $Pb_i = \{T_1,...,T_n\}$, where n is the number of tasks and could be different from one blueprint to another. A task is a function of the capabilities (C) of the robot (R), i.e., $T_i = f(C_x, C_y, ...)$, where each robot owns different capabilities set. If the RqM finds a match between Pb to and a Rq, it forwards the Pb to the PLN. The PLN checks the robots' availability, and their capabilities to achieve the tasks in the Pb. If more than a robot owns the capabilities to fulfill the task, the PLN compares the number of tasks that have been achieved by these robots in the past. Based on this comparison, the PLN selects a robot to assign for the task. If the PLN complete the matching of all the tasks with the robots, it sends a verified plan (Pv) to the RbM. The RbM sends the tasks to the robots and waits their response.

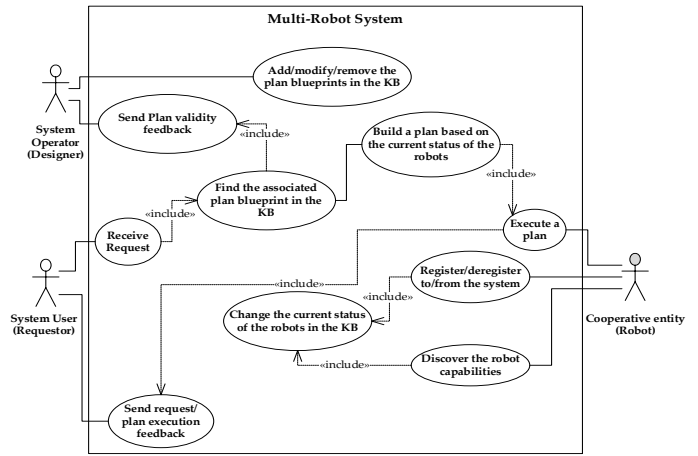

Figure 4: UML/SysML use case diagram for the proposed architecture

The use case diagram in Figure 4 shows three variation types that are considered during the simulation. First is the Pbs variation, by adding, editing, or omitting a Pb. Second is variation in the number of the available robots. The maximum number of robots that can exist is constrained to three. The robots are constrained to register or deregister through the RbM. Third is the variation in the robots' capabilities, by updating or editing the capabilities of a robot. the robot is constrained to deregister to be able to update its capabilities, then register again through the RbM, which automatically updates the robot new capabilities in the KB.

## 3. Solution preliminaries

### 3.1. Systems Modeling Language

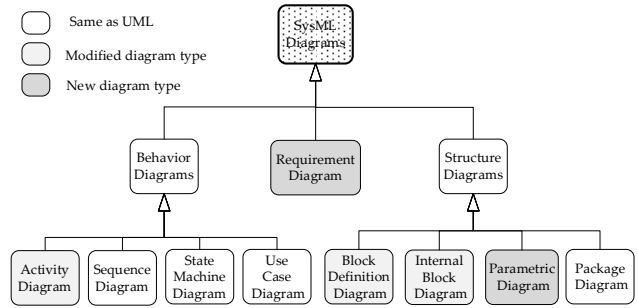

Figure 5: SysML Taxonomy and comparison to UML – adapted from

SysML is a general-purpose modeling language that is derived from Unified Modeling Language (UML) [9]. SysML and UML belong are both developed by Object Management Group (OMG). UML is is a visual modeling language that is particularly used to construct, design, and document the software systems in fields such as web-development, telecommunication, banking, and enterprise services [10]. While SysML is extending and modifying UML diagrams to fit complex industrial systems that involve variety of hardware, software, information, and processes (e.g., Aviation, Space, Automotive) [11].

Figure 5 shows the relation between SysML and UML graphs [12]. Requirement and parametric diagram are two new diagrams that distinguish SysML [13]. Section 4 of this article used the requirement diagram to define the system performance criteria requirements and their relations. Block definition diagram and internal block diagram are used to describe the main components of the system architecture and the connection among the components as illustrated in section 2, while the use case diagrams describe the interaction of the system as a black box with the external world or actors. The activity diagram is used in section 5 to represent the MRS components logic. The more detailed logic is represented in the activity model, the easier to automatically generate a low-level code from this activity model. For this reason, BPMN is used to build the MRS logic, as it extends the notations, semantics, and syntax of SysML and UML activity diagram. The state machine diagram is used in section 4 to model the internal states of the robot. While the sequence diagram is used in section 6 to represent the interaction and communication among the components during a simulation scenario.

*3.2. Business Process Model and Notation*

Since UML activity diagram provides an abstract high-level process description, BPMN extends the UML activity diagram to fulfill the following two drawbacks. First, UML activity diagram lakes the syntax and the logical execution among the actions. Second, the poverty in UML notations and semantics in comparison with BPMN [14].

Table 1: BPMN control gateways

| Rule | Gateways | | |
|---|---|---|---|
| | Exclusive OR (XOR) | Inclusive OR | Parallel AND |
| Split | Decision? one output only can be activated | Decision? more than one output can be activated | Boolean Decision? all outputs can be activated |
| Merge | any of the inputs is active? | more than one input are active? | all of the inputs are active? |

Flow control gateways is the best example to demonstrate how BPMN is improving UML activity diagram. Flow control gateways are all equivalent to only one notation in UML, which is the decision notation. Table 1 shows the notations, semantics, and syntax of the basic gateways of BPMN. Three different notations are demonstrated in Table 1, which are exclusive-OR, inclusive-OR, and parallel-AND. The three mentioned gates operate either as spilt or merge context. In spilt context, exclusive-OR splits one input to only one output based on the conditions on the output branches. Inclusive-OR splits one input to more than one output simultaneously based on the conditions on the output branches. Parallel-AND splits one input to all the output simultaneously when the input branch is triggered. In merge context, exclusive-OR merges any of the input branches to only one output, when any of the input branches is triggered. Inclusive-OR merges more than one input branches to only one output, when these inputs are simultaneously triggered. Parallel-AND merges all the input branches to only one output, when all the inputs are simultaneously triggered [15].

*3.3. Java Agent Development*

JADE is a Multi-Agent System (MAS) middleware [16] that has been used in this research to deploy the proposed solution as shown in Figure 6-a. Each entity in the proposed SysML internal block diagram is implemented as a JADE agent. JADE Agent Management System (AMS) address each agent with a unique Identifier (AID) to facilitate the communication among the agents. While JADE directory Facilitator (DF) announces the services that every agent afford. JADE applies the Foundation for Intelligent Physical Agent (FIPA) specifications, to enable agent communication through FIPA-Agent Communication Language (FIPA-ACL) [17].

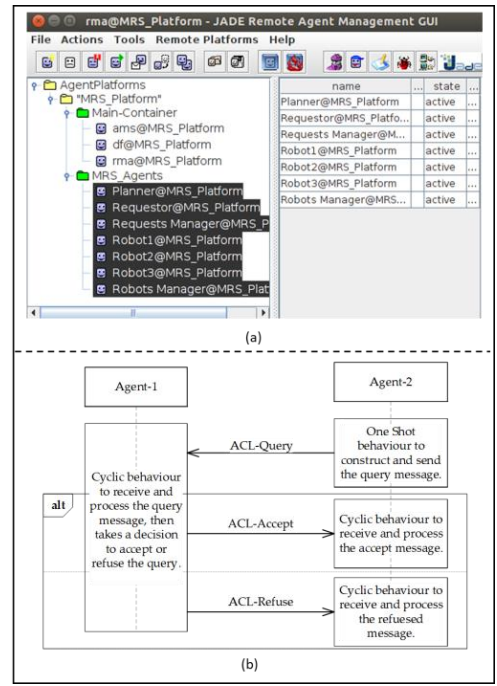

Figure 6 (a) JADE framework – (b) JADE sequence diagram example

Each JADE agent has a complex individual behaviour that can be seen as a composite of two simple behaviours. First is one-shot behaviour that is executed only once when it is triggered. Second is a cyclic behaviour that continuously executed when it is triggered. An example of JADE agent communication and decision making based on their behaviours can be seen in Figure 6-b. JADE is a suitable tool to build an agent simulation based on the MRS SysML/BPMN model. As the MRS logic and architecture can be easily translated to JADE implementation concepts [18].

## 4. Performance requirements

To evaluate the MRS design, it is necessary to measure the system performance during the simulation. Qualitative criteria such as reusability, scalability, extensibility, and interoperability have been proposed in [19]. However, these criteria are often relatively vague without quantitative performance measurements. Therefore, this research defines the quantitative indicators that are shown in Figure 7. The research assumes that the MRS is a black box that receives different Rq, that can either success or fail during the execution. The following measurements can be used to express the system performance:

- Throughput: the number of requests that are processed.
- Latency: the time needed from the request arrival till the request execution.
- Success rate: the number of request that success to be executed per the overall received requests number.
- Failure rate: the number of request that fail to be executed per the overall received requests number.

- Efficiency: the ration between the success rate and the failure rate.

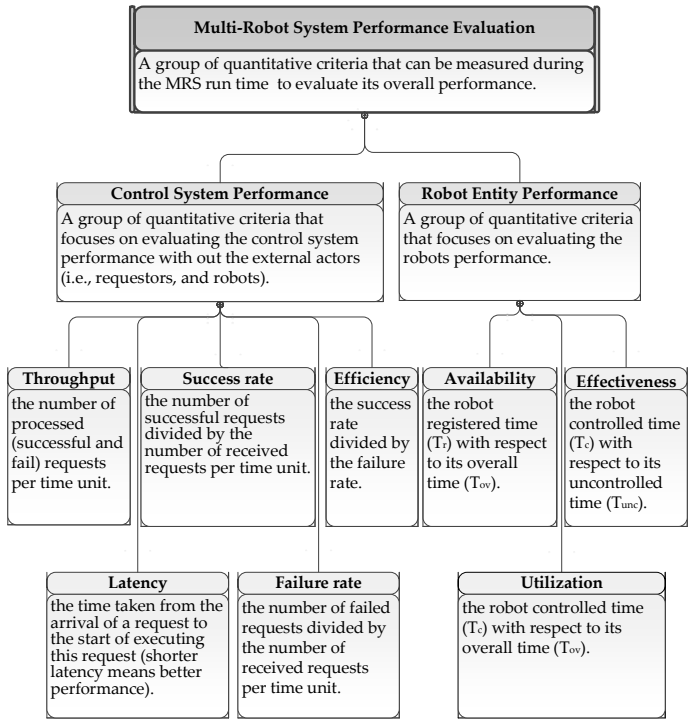

Figure 7: Requirement diagram of the MRS performance evaluation criteria

The robot performance is also considered in this research as another measurement of the MRS performance [20]. The robot performance is fundamentally derived from its state machine diagram that is shown in Figure 8. The the robot state machine is built upon measuring the following times:

- Controlled time ($T_c$): the time that the robot needs to perform an assigned task.
- Uncontrolled time ($T_{unc}$): the robot waiting time to be assigned to a task after registration.
- Registered time ($T_r$): the sum of the controlled and the uncontrolled time of the robot.
- Unregistered time ($T_{unr}$): the accumulation of the robot unregistered time.
- Overall time ($T_{ov}$): the sum of the registered and the unregistered time of the robot.

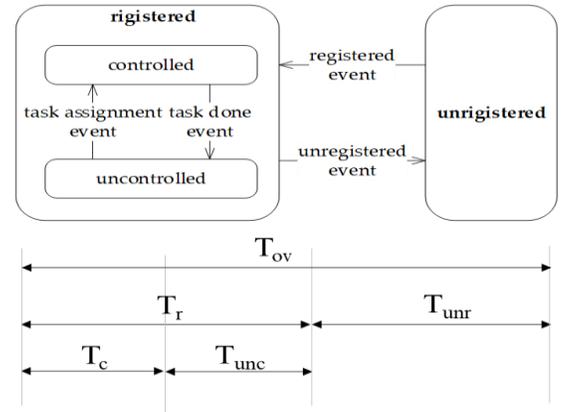

Figure 8: State machine diagram of the robot entity

Accordingly, the robot performance criteria are calculated as follows:

- Availability: the ration between the robot registered time ($T_r$) and the overall time ($T_{ov}$).
- Utilization: the ratio between the robot controlled time ($T_c$) and the overall time ($T_{ov}$).
- Effectiveness: the ration between the robot controlled time ($T_c$) and the uncontrolled time ($T_{unc}$).

## 5. System model

### 5.1. Requests manager

The RqM receives requests from various requestors, then it looks for an associated Pb within the KB. If the RqM finds the associated Pb, it forwards it to the PLN. The RqM decision making model is shown in Figure 9 via the BPMN activity diagram.

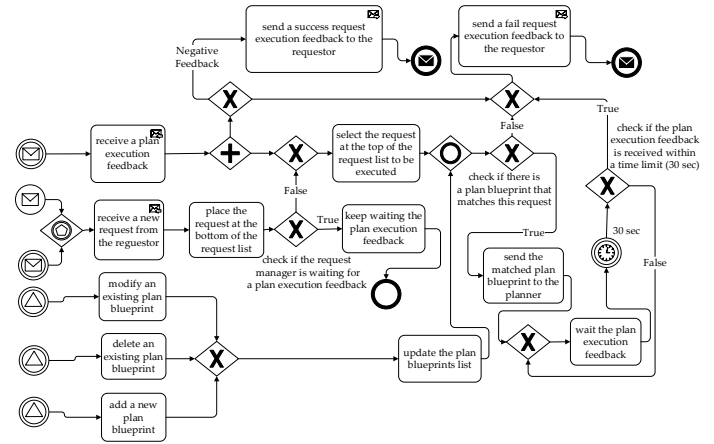

Figure 9: Requests-manager BPMN activity diagram

The RqM uses First Come First Serve (FCFS) technique to schedule the received requests. The RqM checks in the associated Pb for every received request. If there is no associated Pb with the request, the RqM directly sends a negative feedback to the requestor. If the RqM finds an associated Pb to the request, it forwards this Pb to the PLN, and waits for the feedback. If this feedback exceeds predefined limits, the RqM considers this request as a failure one. If not, it waits the execution feedback to forward it to the requestor.

### 5.2. Planner

The PLN receives the Pb and makes sure that it is visible to build a Pv instance according to the current system status. The PLN decision making model is shown in Figure 10 via the BPMN activity diagram.

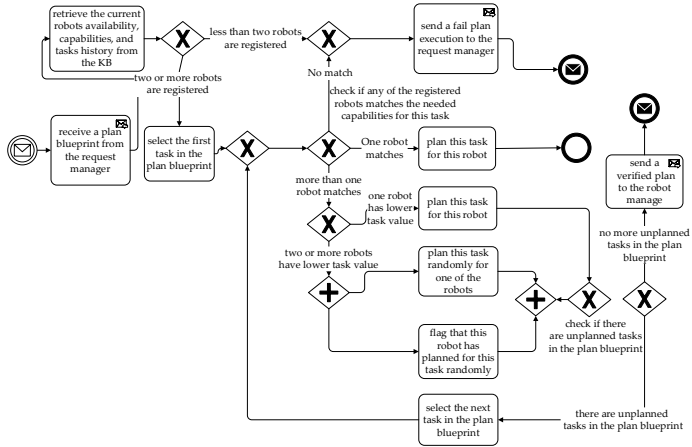

Figure 10: Planner BPMN activity diagram

To construct a Pv instance from a Pb, The PLN checks the available registered robots, the robots' capabilities, and the robots' tasks history. In case that there is only one available robot, the PLN directly considers a plan failure, as it is known in advance that a plan requires at least two robots to get executed. If at least two robots are available, the PLN compares the tasks in the Pb to the available robots' capabilities. If the robots' capabilities do not match the required tasks in the Pb, the PLN considers a plan failure. If there are two robots or more that can perform the same task, the PLN checks their tasks history, and assign the task to the robot that performed less tasks. This is to balance the task assignment among the available robots within the MRS. If all the tasks in the Pb could be assigned to robots, the PLN creates a Pv instance and sends it is the RbM to be executed.

*5.3. Robots Manager*

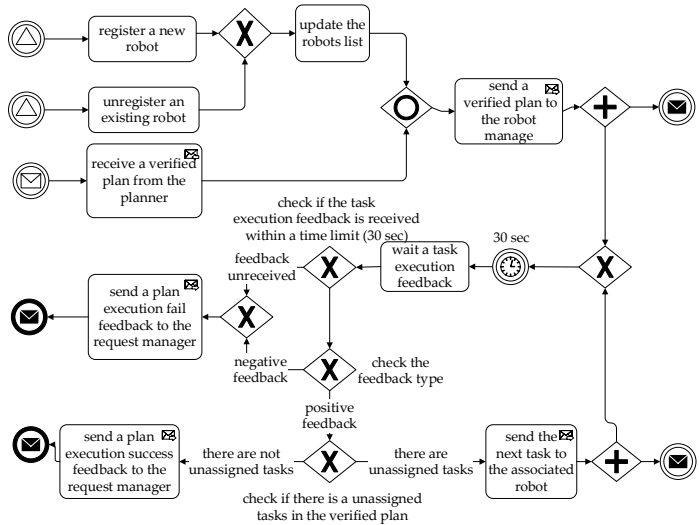

Figure 11: Robots-manager BPMN activity diagram

The RbM receives the Pv, then it assigns the tasks in this Pv to the available robot. Additionally, the RbM is also responsible for registering/unregister the robots from the MRS. this way it monitors the robots' availability. The RbM decision making model is shown in Figure 11 via the BPMN activity diagram.

When the RbM assigns a task to a robot, it waits the robot feedback within a time limit. If the robot feedback did not arrive within the predefined limits, the RbM sends a negative feedback to the RqM. This feedback means that the whole plan is failed to be executed. If the RbM received a positive feedback from the robot within the predefined time limits, it assigns the next task due to the Pv. If all the tasks in the Pv are executed, the RqM sends a positive feedback to the RqM, otherwise it sends a negative feedback.

## 6. Simulation

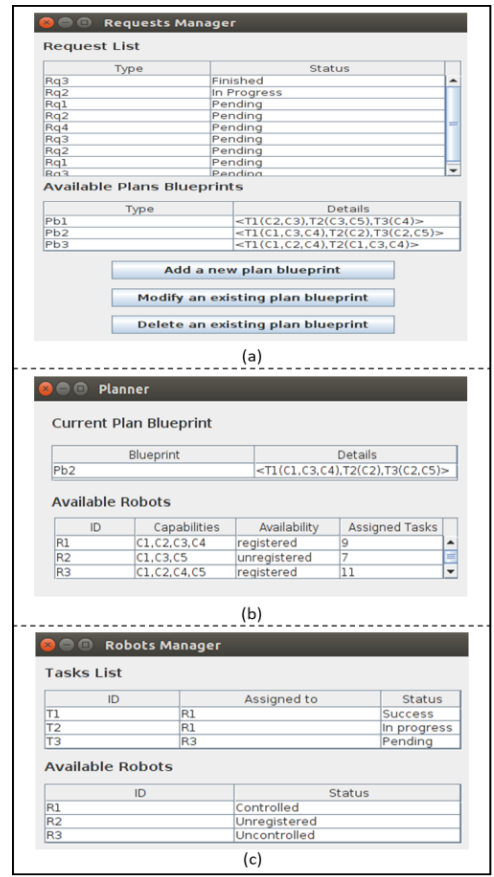

Figure 12: (a) Requests Manager GUI – (b) Planner GUI – (c) Robots Manager GUI

The activity diagrams that have been illustrated in the previous section are used as the MRS blueprints. JADE has been used in this research to deploy these blueprints, and hence enables the MRS simulation during the design phase. The Graphical User Interface (GUI) shown in Figure 12 has been created to achieve interact with every entity in the proposed architecture. The RqM GUI in Figure 12-a can be used to add/edit/remove the Pb. The PLN GUI in Figure 12-b is used to monitor the Pv execution, the robots' availability, the robots' status, the robots' capabilities, and the robots' tasks history. The RbM GUI in Figure 12-b is used to show the assigned tasks status.

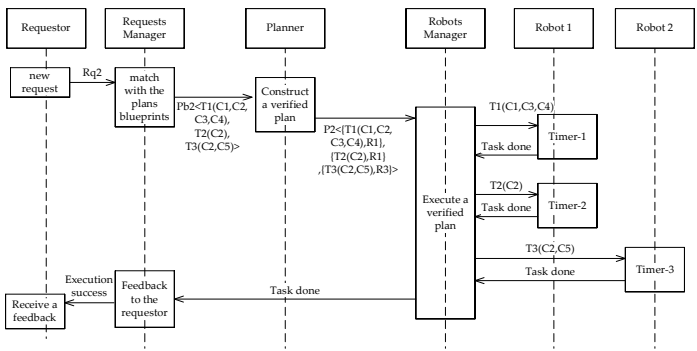

Figure 13: Simulation scenario plan execution sequence diagram – an example

To illustrate the simulation scenario, an interaction example among the MRS entities is show in Figure 13. In this example, The RqM receives $Rq_2$. Therefore, the RqM sends the Pb in a form of the ACL-message shown in Figure 14-a to the PLN. Accordingly, the PLN constructs a Pv by matching the available robots' capabilities and tasks history with the received Pb. In this case, $R_1$ and $R_3$ were registered into the MRS as shown in Figure 14-b. As $T_1$ needs ($C_1$, $C_3$, $C_4$) to be executed, $T_1$ was assigned to $R_1$, because ($C_1$, $C_3$, $C_4$) are unique capabilities of $R_1$. Similarly, $T_3$ was assigned to $R_3$, as $T_3$ needs ($C_2$, $C_5$) which is unique capability of $R_3$. However, in case of $T_2$, both $R_1$ and $R_3$ own the capability $C_2$ which is needed to execute this task. Therefore, the PLN checks both robots' task history to be able to assign $T_2$. The PLN finds out that $R_1$ task history is 9 while $R_3$ task history is 11. Accordingly, the PLN assigns $T_2$ to $R_1$, to balance the robots' tasks distribution.

Ultimately, the PLN sends the Pv in form of the ACL-message shown in Figure 14-b to the RbM. The RbM assigns the tasks to the associated robots according to the Pv. The task assignment is sent as an ACL-message as shown in Figure 14-c. The RbM waits the robots' feedback within a timeframe window. If all the RbM received success feedbacks for all the assigned tasks, it sends a plan success feedback to the RqM.

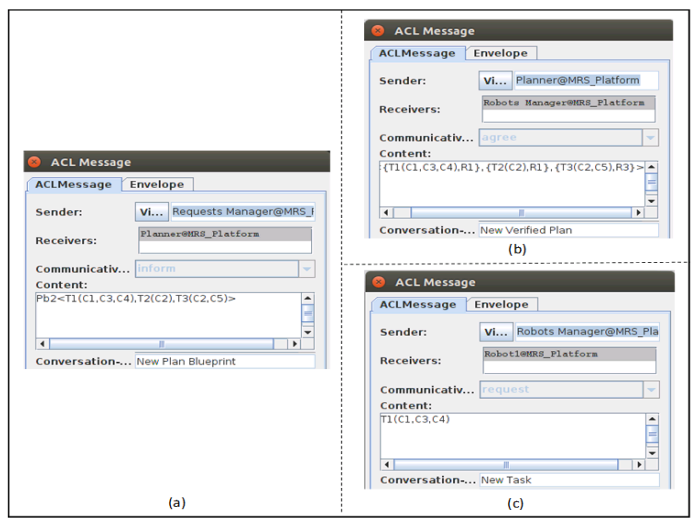

Figure 14: (a) Plan blueprint message – (b) Verified Plan message – (c) Assigned task message

## 7. Simulation results analysis

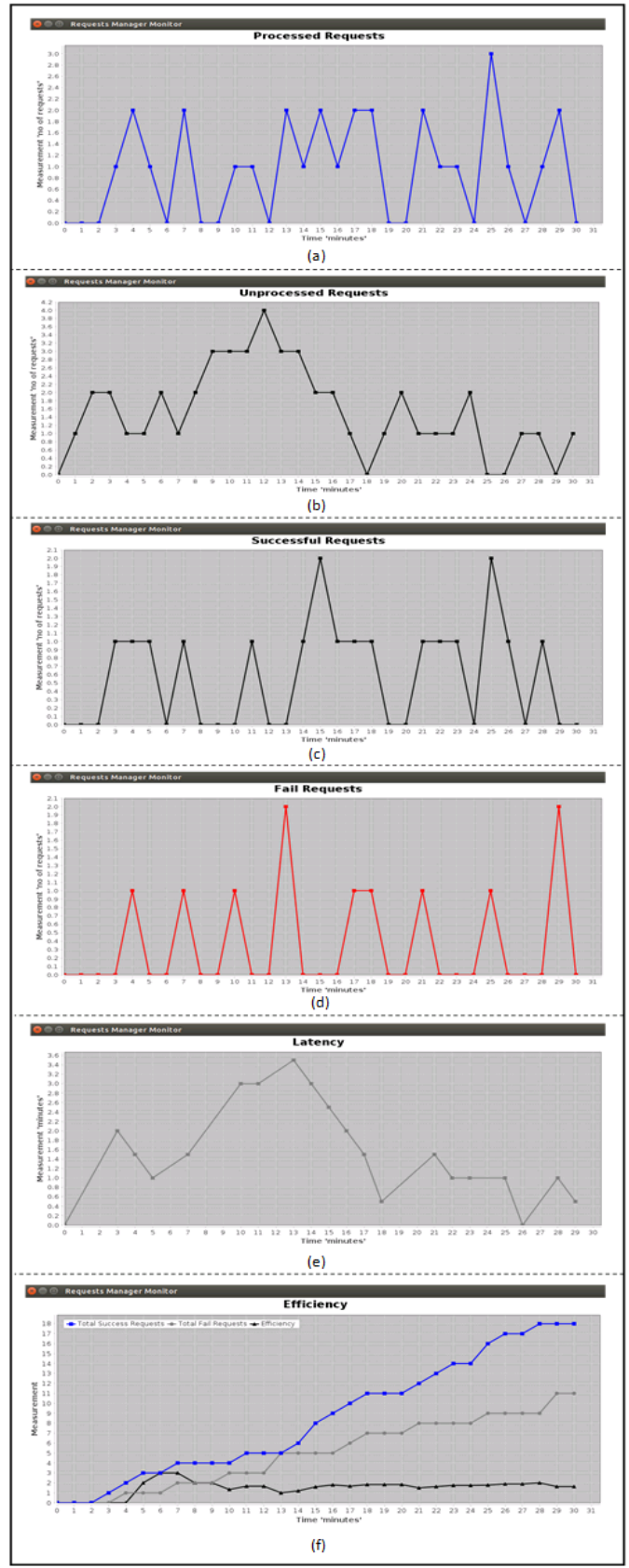

Figure 15: Simulation graphs (a) Processed requests – (b) Unprocessed request – (c) Successful requests – (d) Failed requests – (e) Latency – (d) Efficiency

As it has been demonstrated in the previous section, the robots' availability, the robot's capabilities, and the the plan blueprints are the variables that can be used to build different simulation scenarios. Accordingly, to measure the system performance, the robots' availability was randomly altered during the run time. Thus, analyzing the simulation results has been done by running JADE MAS for 30 minutes as shown in Figure 15, then measuring the system performance indicators that are concluded in section 4. Each one minute, a new request is generated, one robot randomly unregister from JADE MAS, and one random robot register to JADE MAS. the robot's capabilities and the the plan blueprints do not change during the simulation scenario.

One of the RqM responsibilities is to monitor the requests status. The number of processed requests by the RqM is shown in the graph in Figure 15-a. Accordingly, the MRS throughput can be directly calculated from this chart. On the one hand, MRS throughput expresses how fast the system, therefore it is a relative value. Thus, to understand the MRS throughput, Figure 15-c and Figure 15-d should be considered as well. For instance, the number of requests at minute 4 is two requests as can be seen in Figure 15-a. But, if we look closely into Figure 15-c and Figure 15-d, we will find out that one request is success and another fail. This means that, it is not important if the system is so fast, but most of the requests are failed to be executed. On the other hand, MRS latency expresses how much delay in the system as it can be seen in Figure 15-e. If the system delay value is equal to zero as can be seen in the 26th minutes of Figure 15-e, this means that the number of unprocessed requests is equal to zero as well, as can be seen in the 26th minutes of Figure 15-b.

The MRS efficiency graph shown in Figure 15-f is derived from dividing the data in Figure 15-c (successful requests) by the data Figure 15-d (fail requests). The MRS efficiency value is absolute. When the MRS efficiency is higher than one, this means that the number of success requests is higher than the number of fail request. Figure 15-f shows that the simulated MRS efficiency is higher than or equal to one during the simulation runtime.

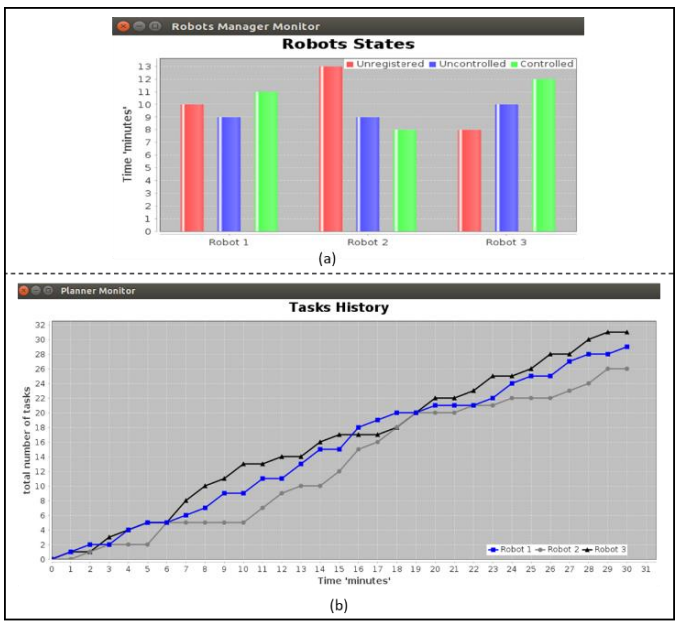

Figure 16: (a) Robots States – (b) Robots tasks history

Table 2: Robots availability, utilization, and effectiveness

|  | Robot 1 (R1) | Robot 2 (R2) | Robot 3 (R3) |
|---|---|---|---|
| Availability $\frac{T_r}{T_{ov}}$ | $\frac{20\ min}{30\ min} = 0.67$ | $\frac{17\ min}{30\ min} = 0.57$ | $\frac{22\ min}{30\ min} = 0.73$ |
| Utilization $\frac{T_c}{T_{ov}}$ | $\frac{11\ min}{30\ min} = 0.37$ | $\frac{8\ min}{30\ min} = 0.27$ | $\frac{12\ min}{30\ min} = 0.4$ |
| Effectiveness $\frac{T_c}{T_{unc}}$ | $\frac{11\ min}{10\ min} = 1.1$ | $\frac{8\ min}{9\ min} = 0.89$ | $\frac{12\ min}{10\ min} = 1.2$ |

One of the PLN responsibilities is to monitor the balance the tasks among the available robots. Figure 16-a shows that the robots' available is changing over the simulation runtime. Simultaneously, Figure 16-b shows that the PLN compensates this variation by balancing the MRS. For instance, the task distribution among the available robot is converging to be 6 tasks per robot at the 6th minute of the simulation. Then, the robots' tasks distribution is diverging till it balanced again to be 20 tasks per robot at the 19th minute of the simulation. Table 2 can be also concluded from Figure 16-a. In this table, $R_3$ is the most utilized and available robot during the simulation runtime, and hence $R_3$ is the most effective in comparison to $R_1$ and $R_2$. Accordingly, the PLN compensates this variation by maximizing $R_1$ and $R_2$ task assignment, to balance them with $R_3$.

## 8. Summary and Discussion

This article has highlighted new dimensions of the MRS design problem, which are the formalization, simulation, and evaluation of the solution architecture. The proposed modeling approach is based on a formal generic ADLs, that can be used to transfer the solution concept over different system case studies, regardless the implementation technology. Furthermore, the illustrated simulation method can be used to verify different architecture design patterns, based on the concluded system performance measurements.

The fundamental SysML diagrams have been implemented to design the proposed MRS system model. Moreover, BPMN language has been used to implement the activity diagram as it extends UML/SysML notations, semantics, and syntax. The collection of these standard models is used as the MRS blueprints. Those blueprints can be easily coded in any programming environment that supports distributed system implementation. For instance, JADE has been used in this research to implement these blueprints, however Robot Operation System (ROS) or Web Service (WS) are very suitable candidates to deploy the system.

A group of MRS performance requirements have been defined during this article, to quantify the system performance during the simulation runtime. Those criteria can are technology agonistic as well, which means that they can be used to compare between the system performance when it is implemented with different technologies. Furthermore, the system simulation is not only used during the design phased, but it can be reused in a form of a real time digital twin during the implementation phase. For instance, to check in advance different planning and scheduling algorithms before executing them on the real system.

Using a formal description language such as SysML or BPMN enables separating the model from the code, which is a common domain specific programming method. Therefore, in the future work, we will write a code generator that can be used to automatically generate the implementation code. Therefore, the model that has been developed in this article will turn to be executable and will be used as the main software artifact of the project. This can dramatically reduce the coding time and effort and improve the system readability and maintainability. Additionally, in the future work, the same performance measurements that have been used in this article can be used in the implementation phase, as a part of the system visualization.